\pdfoutput=1

\documentclass[11pt]{article}

\usepackage{acl}

\usepackage{times}
\usepackage{latexsym}

\usepackage[T1]{fontenc}

\usepackage[utf8]{inputenc}

\usepackage{microtype}

\usepackage{inconsolata}

\usepackage{graphicx}

\usepackage{booktabs}
\usepackage{makecell}    
\usepackage{cleveref}
\usepackage{amssymb}
\usepackage{listings}
\usepackage{csquotes}
\usepackage{mdframed}

\usepackage{soul}

\crefname{lstlisting}{listing}{listings}
\Crefname{lstlisting}{Listing}{Listings}

\usepackage{todonotes}
\definecolor{rot}   {RGB}{204   7  30}

\title{Listen to the Context: Towards Faithful Large Language Models\\for Retrieval Augmented Generation on Climate Questions}

\author{David Thulke\textsuperscript{1,2} \, Jakob Kemmler\textsuperscript{1,2} \, Christian Dugast\textsuperscript{2} \, Hermann Ney\textsuperscript{1,2} \\ \\
        \textsuperscript{1}Machine Learning and Human Language Technology, RWTH Aachen University, Germany \\
        \textsuperscript{2}AppTek GmbH, Aachen, Germany\\
        \texttt{\{thulke,jakob.kemmler,ney\}@hltpr.rwth-aachen.de, cdugast@apptek.com}}

\begin{document}
\maketitle
\begin{abstract}
Large language models that use retrieval augmented generation have the potential to unlock valuable knowledge for researchers, policymakers, and the public by making long and technical climate-related documents more accessible.
While this approach can help alleviate factual hallucinations by relying on retrieved passages as additional context, its effectiveness depends on whether the model's output remains faithful to these passages.
To address this, we explore the automatic assessment of faithfulness of different models in this setting.
We then focus on ClimateGPT, a large language model specialised in climate science, to examine which factors in its instruction fine-tuning impact the model's faithfulness.
By excluding unfaithful subsets of the model's training data, we develop ClimateGPT Faithful+, which achieves an improvement in faithfulness from 30\% to 57\% in supported atomic claims according to our automatic metric.
\end{abstract}

\section{Introduction}

As the urgency of climate action intensifies \cite{le06900s}, researchers, policymakers, and the public require efficient access to reliable climate information.
Large language models (LLMs) have emerged as powerful tools for processing and summarizing vast amounts of climate-related literature.
However, these models often suffer from factual hallucinations, where generated responses contain incorrect or misleading information.
One promising approach to mitigating this issue is Retrieval Augmented Generation (RAG), which enables models to ground their responses in retrieved passages from trusted sources.
Several climate-related RAG-based systems have been developed, including ChatClimate \cite{chatclimateVaghefi2023}, ClimateGPT \cite{thulke2024climategptaisynthesizinginterdisciplinary}, and ClimateQ\&A \cite{calzada2024climateqa}.
Despite these advancements, challenges remain in ensuring that generated outputs remain faithful to the retrieved content.

\begin{figure}[b!]
        \begin{mdframed}
            \fontsize{10pt}{12pt}\selectfont
        \fontsize{10pt}{12pt}\selectfont \textbf{Prompt}

        {What is the purpose of the Global Stocktake?} \newline

        \fontsize{10pt}{12pt}\selectfont \textbf{Context}

        {%
        [0] Recalling Article 14 of the Paris Agreement, the objective of the Global Stocktake is to assess the collective progress towards achieving [...] \newline
        [1] · Paragraph 34 of 19/CMA.1 decides that the outputs of this component of the global stocktake should: a) Identify opportunities for and [...]}\newline
        [2] The SBSTA and the SBI invited submissions on views on the approach to the consideration of outputs component of the first global [...]\newline
        
        \fontsize{10pt}{12pt}\selectfont \textbf{ClimateGPT 7B}  

        {%
        The Global Stocktake is a key component of the Paris Agreement, \textcolor{rot}{scheduled to begin in 2023}. It involves a comprehensive assessment of global progress in addressing climate change, \textcolor{rot}{considering scientific evidence, economic implications, and societal impacts}. The Stocktake aims to inform the updates to countries' climate commitments, \textcolor{rot}{known as \enquote{nationally determined contributions} (NDCs)}, \textcolor{rot}{to ensure they are in line with the agreement's objectives and targets}. [...]} \newline

        \fontsize{10pt}{12pt}\selectfont \textbf{ClimateGPT 7B Faithful+}

        {%
        The Global Stocktake's purpose, as outlined in Article 14 of the Paris Agreement, is to assess the collective progress made towards achieving the Agreement's purpose and long-term goals, which encompass mitigation, adaptation, and means of implementation. This assessment is conducted considering equity and the best available science. The Global Stocktake process spans two years, beginning with information collection and preparation and culminating in a consideration of [...]}
        \end{mdframed}
    \caption{Shortened example from the original ClimateGPT and the Faithful+ variant on one example from the Climate Policy Radar task. Text marked in \textcolor{rot}{red} is not faithful, i.e. it is not supported by the given context. The full example is shown in \Cref{fig:stocktake-example}.}
    \label{fig:intro-example}
\end{figure}

Faithfulness, in this context, refers to the extent to which a model's response accurately reflects the information contained in the retrieved passages without introducing extraneous or misleading details.
Importantly, factuality does not imply faithfulness.
A response may be factually correct with respect to general world knowledge but still unfaithful if the information is not supported by the retrieved passages as shown in \Cref{fig:intro-example}.
A lack of faithfulness undermines trust in these models, particularly in the climate domain, where misinformation has significant real-world consequences.

Moreover, we argue that faithfulness is even more important than general factuality in this setting, as large language models may inevitably hallucinate when faced with long-tail or rare knowledge.
By requiring that all factual information in a response originates from the provided context, we can mitigate the risk of such hallucinations and ensure that model outputs are transparent, verifiable, and aligned with the available evidence.
Thus, evaluating and improving faithfulness is a crucial step in enhancing the reliability of climate-focused LLMs.

In this work, we investigate methods for automatically assessing the faithfulness of RAG-based models in climate-related applications.
We then focus on ClimateGPT \cite{thulke2024climategptaisynthesizinginterdisciplinary}, a specialised open-weight LLM trained on climate-related texts to examine how different instruction fine-tuning (IFT) datasets influence faithfulness.
By excluding parts of the training data with low faithfulness, we propose a new model ClimateGPT Faithful+ that on our main benchmark increases the percentage of supported claims from 30\% to 57\%.

\begin{table*}[t]
    \centering
    \fontsize{10pt}{12pt}{
    \begin{tabular}{lrrcrrr}
        \toprule
                        & \textbf{\#Tokens} & \textbf{\#Parameters} &              & \textbf{Avg.}     & \multicolumn{2}{c}{\textbf{Claim Support wrt.}} \\
         \textbf{Model} & \textbf{in Trillion} & \textbf{in Billion}   & \textbf{RAG} & \textbf{\#Claims} & \textbf{Ref. \small{[\%]}} & \textbf{KB \small{[\%]}} \\
        \midrule
        LLama 3.1 Instruct & 15 & 8  & -  & 22.7 & -                     & 59 \\
                           &  &  & \checkmark & 17.3 & \fontsize{11pt}{12pt}{\bf{67}}                     & \fontsize{11pt}{12pt}{\bf{72}} \\
        \midrule
            LLama 2 Chat & 2 &7  & -  & 23.3 & -                    & 60 \\
                         &  &   & \checkmark & 21.2 & 48                   & 65 \\
        \midrule
        ClimateGPT & 2&7  & -  & 21.6 & -                     & 59 \\
                   &  &  & \checkmark & 21.1 & 30                     & 61 \\
        \midrule
        ClimateGPT Faithful+ (ours) & 2&7 & -  & 20.2 & -                     & 57 \\
                                  &  & & \checkmark & 19.2 & \ul{57}                     & \ul{69} \\
        \bottomrule
    \end{tabular}}
    \caption{Results for claim support wrt.\ the reference, as a metric of faithfulness, and wrt.\ the knowledge base (KB) as a metric for factuality for different large language models with and without RAG.}
    \label{tab:overall_results}
\end{table*}

\section{Faithfulness and Factuality}

Our definition of faithfulness and factuality follows the work of 
\citet{dziri-etal-2022-faithdial} and \citet{lei2025hallucinationsurvey}.
Given a question $q$, a set of $N$ retrieved passages $K = (k_1, k_2, \ldots, k_N)$ from a knowledge base $\mathcal{KB}$, and a response $r$, we define faithfulness of $r$ with respect to $K$ as $r$ should be supported by the information in $K$, i.e., $r$ should not contain any information that contradicts the information in $K$ or is not present in $K$.
Factuality, on the other hand, refers to the correctness of the information in $r$ with respect to general world knowledge.
In our context, we assume that the relevant world knowledge is contained in $\mathcal{KB}$.
Thus, we consider a response $r$ to be factual if it is faithful to $\mathcal{KB}$.

\subsection{Evaluation}
\label{sec:eval_faithfulness}

To assess both the faithfulness and factuality of long-form responses, we build upon existing automated evaluation approaches, particularly RAGAs \cite{es-etal-2024-ragas} for faithfulness and FActScore \cite{min-etal-2023-factscore} and VeriScore \cite{song-etal-2024-veriscore} for factuality.
These methods share a common three-step pipeline: (1) claim decomposition, (2) evidence retrieval, and (3) claim verification.
The main differences in evaluating for faithfulness versus factuality lie in the evidence retrieval step, as we describe below.

\textbf{Claim Decomposition}\,
As long-form responses are typically composed of multiple claims, we first decompose the response into smaller and independent claims to simplify the subsequent steps.
Given a response $r$, we decompose it into a set of claims $C = {c_1, \dots, c_I}$.
The definition of a claim and the granularity of the decomposition differs between different variants and use-cases.
In this work, we use the claim decomposition method from RAGAs \cite{es-etal-2024-ragas} which prompts a large language model to decompose the full response into smaller claims in one step.

\textbf{Evidence Retrieval}\,
The key distinction between evaluating faithfulness and factuality lies in this step.
For faithfulness evaluation, we directly use the retrieved passages $K = {k_1, k_2, \dots, k_N}$ from the RAG process as evidence.
In contrast, for factuality evaluation, relevant evidence for each claim $c_i$ is retrieved from a knowledge base $\mathcal{KB}$.
In this work, we use the retrieval mechanism that is also used for RAG.

\textbf{Claim Verification}\,
Finally, for each claim, we verify whether it is supported by the retrieved evidence.
Therefore, we use an LLM to classify each claim $c_i$ given the retrieved evidence (multiple retrieved evidence passages are concatenated into a single evidence).
Similar to other work \cite{song-etal-2024-veriscore}, we do not differentiate between refuting and unrelated evidence.
The overall faithfulness and factuality scores of $r$ are then aggregated from these individual claim verifications by reporting the percentage of supported claims.

\textbf{Implementation Details}\,
The exact prompts we used for each step are reported in \Cref{appendix:ragas_prompts}.
GPT-4o (version \texttt{gpt-4o-2024-08-06}) is used as the large language model.

\section{ClimateGPT IFT Evaluation Task}

We use the same evaluation dataset and RAG setup as \citet{thulke2024climategptaisynthesizinginterdisciplinary} to evaluate the faithfulness and factuality of the generated responses.
The test set is a held-out portion of the IFT data curated to train ClimateGPT.
It was created in cooperation with domain experts and contains different open-ended tasks like QA, text generation, classification, chat, and brainstorming as well as closed-ended tasks like summarisation, extraction or rewrite.
Our evaluation focuses on the subset of open-ended prompts of the held-out data (334 out of the 400 samples).

\subsection{Information Retrieval}

We use the dataset and retrieval pipeline as described by \citet{thulke2024climategptaisynthesizinginterdisciplinary} for retrieving relevant contexts in our faithfulness evaluation.
The dataset consists of climate-related documents from various sources, including IPCC reports and climate science related papers (see \Cref{appendix:retrieval_dataset} for detailed statistics).
For retrieval, we employ the \texttt{bge-large-en-v1.5} embedding model \cite{xiao2024bge} and a hierarchical retrieval strategy where we first retrieve the most relevant pages based on the query, selecting the top 5 ranked pages.
Then, within these, we retrieve the top 5 most relevant 115-token snippets.

\subsection{Large Language Models}

We experiment with several language models in addition to ClimateGPT.
As baselines, we include the 7B parameter variants of Llama 2 Chat (which shares the same foundation model as ClimateGPT) and Llama 3.1 Instruct.
Further, we report results on the 70B parameters variants as well as on GPT-4o in \Cref{tab:overall_results_all_models} in the appendix.
For all baseline models, we use a standardized RAG prompt that explicitly instructs the model to base its response solely on the provided references\footnote{Full prompt in \Cref{appendix:rag_prompt}.}.
Both the user question and retrieved references are included within the user message to ensure a consistent evaluation setup.
For ClimateGPT, we leverage its dedicated context role, which was introduced during training to optimize reference usage.
We also use the model's default system prompt to align with its intended deployment configuration.

\begin{table*}
    \centering
    \fontsize{10pt}{12pt}{
    \begin{tabular}{llrrrr}
        \toprule
         & & & \textbf{Avg.} & \textbf{Claim Support wrt.} \\
        \textbf{Source} & \textbf{Subset} & \textbf{Size} & \textbf{\#Claims} & \textbf{Ref.\ \small{[\%]}} \\
        \midrule
        Senior Expert & Grounded & 74  & 8.6 & 93 \\
        Expert & Grounded & 403  & 13.1 & 52 \\
        Non-Expert & Open-Ended & 8,503 & 19.1 & - \\
         & Closed-Ended & 1,160  & 10.0 & 90 \\
         & (Open-Ended) Grounded & 2,368  & 19.0 & 43 \\
         & (Closed-Ended) Grounded & 1,024  & 9.6 & 91 \\
        \bottomrule
    \end{tabular}}
    \caption{Climate-specific subsets of the ClimateGPT IFT data. For the closed-ended examples, claim support wrt.\ reference refers to the context given in the prompt and for grounded examples it refers to the given paragraphs.}
    \label{tab:ift_table}
\end{table*}

\subsection{Results}

We report the results with our faithfulness and factuality metrics for the small models in \Cref{tab:overall_results}. Results of all models are reported in \Cref{appendix:full_results}.
Overall, we observe that the more recent Llama 3.1 has significantly higher faithfulness than the predecessor Llama 2.
For ClimateGPT, we observe that the faithfulness, as measured by claim support, is very low. 
Further, in contrast to the other models, using RAG with ClimateGPT does only slightly improve the claim support wrt. to the KB, i.e.\ the factuality.
This is a strong indicator that the model does not make effective use of the provided paragraphs.

Factuality, i.e.\ claim support in the knowledge base might be underestimated.
By looking at claims that are not supported by the knowledge base, we identify multiple instances of claims that are factual but where we fail to retrieve the relevant evidence.
This either occurs due to the limited size of our knowledge base or due to a failure on retrieval.
For an assessment of factuality, we therefore note that the reported metric should just be considered as a lower bound and more accurate results could be achieved.

\section{Ablation of the IFT Data}

\begin{table*}
    \centering
    \fontsize{10pt}{12pt}{
    \begin{tabular}{lccccrr}
        \toprule
        & \textbf{Other}
        & {\textbf{Open-End.}}
        & {\textbf{Closed-End.}}
        & {\textbf{Grounded}}
        & \textbf{Avg.} 
        & \textbf{Claim Support} \\
        \cmidrule(lr){1-5}
        \textbf{Size} & 65,000 & 8,503 & 1,160 & 3,328 & \textbf{\#Claims} & \textbf{wrt. Ref \small{[\%]}} \\
        \midrule
        ClimateGPT 7B& \checkmark & \checkmark & \checkmark & \checkmark & 21.1 & 30 \\
        \cmidrule(lr){2-7}
        & \checkmark & \checkmark & \checkmark & - & 19.2 & \fontsize{11pt}{12pt}{\bf{57}} \\
        & \checkmark & \checkmark & - & - & 18.9 & 49 \\
        & \checkmark & - & \checkmark & - & 20.1 & \fontsize{11pt}{12pt}{\bf{58}} \\
        & \checkmark & - & - & - & 20.4 & 53 \\
        \bottomrule
    \end{tabular}}
    \caption{Ablation study results showing test-time claim support for different training data combinations.}
    \label{tab:ablations_table}
\end{table*}

Motivated by the suboptimal faithfulness of ClimateGPT, especially compared to Llama 2 Chat, we want to study the post-training of the model.
We focus on the IFT step as we do not expect that the continued pre-training step has a significant impact on the faithfulness of the model.
The IFT data of ClimateGPT consists of 
a general domain partition and 
a climate-specific partition that was specifically curated to train the model.
The different subsets of the latter are listed in \Cref{tab:ift_table}.
A small portion of the data was generated in close cooperation with domain experts (Exp.), and the larger set generated by non-experts (Non-Exp.).
In \emph{closed-ended} questions, the model is given a reference text to perform its task, such as creating a summary of that text or extracting specific information from it.
In contrast, for \emph{open-ended} questions, no additional explicit references are given in the prompt, and the model is expected to use its parametric knowledge or to retrieve additional sources via RAG.

\textit{Grounded} refers to examples where additional context is provided to the model as it would be the case when RAG is used during inference.
For the expert and senior expert subsets, these references were directly provided during annotation.
In the case of the non-expert subset, annotators only provided one or multiple URLs to sources the answer is based on.
For a subset of the dataset, these URLs were crawled, chunked and \citet{thulke2024climategptaisynthesizinginterdisciplinary} used a heuristic\footnote{See Section 4.3 in \citet{thulke2024climategptaisynthesizinginterdisciplinary} for more details.} to select the most relevant chunk as context for the response.
Additionally, for each example up to four distractor paragraphs from other documents were selected to make the model more robust to noisy retrieval results.
For closed-ended questions, only distractors were added as all the relevant content is already provided in the prompt.

We start our investigation by analysing the faithfulness of the gold responses in the IFT data with respect to their context.
For closed-ended questions, we use the full prompt as context and for the grounded questions, the selected context paragraphs.
The percentage of supported claims for each subset as well as the average number of claims per response are reported in \Cref{tab:ift_table}.
We notice that the Grounded Senior Expert and Closed-Ended Non-Expert are faithful to their context with 93\% and 90\% of claims being supported.
The faithfulness of the Grounded Expert data is already much lower with only 52\% claim support.
Upon closer inspection, we found that the annotators only provided grounding passages for crucial claims in the response.
Finally, we observe the lowest level of faithfulness for the Open-Ended Grounded Non-Expert data with only 43\% claim support.

Next, we repeated the IFT step on different subsets of the data to observe the effect on the faithfulness on the final model.
The results are reported in \Cref{tab:ablations_table}.
As anticipated from our previous analysis of the IFT subsets, excluding the grounded non-expert data significantly increases the claim support from 30\% to 57\%.
Furthermore, excluding the closed-ended but not grounded non-expert data reduces the claim support again to 49\%.
This indicates that closed-ended examples with high faithfulness seem to improve the faithfulness of the model despite the context being provided directly in the prompt, rather than via retrieval.
Finally, removing the open-ended non-expert data without grounding does not have a significant effect on the claim support.
For the final model, \emph{ClimateGPT Faithful+} we still include the open-ended data as we expect it to improve other aspects.

\subsection{Validation on Alternate Metrics and Tasks}

To validate the generalizability and robustness of our improvements, we conducted additional experiments on a RAG dataset from Climate Policy Radar \cite{juhasz2024responsibleretrievalaugmentedgeneration} focussing on questions on climate policy documents.
For ClimateGPT Faithful+, we observe a similar improvement in faithfulness with an improvement in claim support from 44\% to 58\%. More details are discussed in \Cref{sec:cpr}.

Further, we confirm the results on the ClimateGPT IFT Task by using an additional faithfulness metric (LettuceDetect,  \citet{kovács2025lettucedetecthallucinationdetectionframework}) and observe an improvement from 6\% to 34\% completely faithful responses with the Faithful+. Details are discussed in \Cref{appendix:results_lettuce_detect}.

\section{Related Work}
\label{sec:related}

Similar to our work, \citet{schimanski-etal-2024-towards} study the faithfulness of a RAG system on climate questions.
They restrict the output of the model so that one sentence always corresponds to exactly one reference passage and verify the faithfulness using an NLI model.
This way they avoid the claim decomposition step. They also fine-tune the model on a synthetic dataset following these constraints to improve faithfulness.
Our work focuses on improving faithfulness by fine-tuning on more complex human written responses.

In addition to the faithfulness evaluation approaches discussed in this work, there are other approaches to evaluate faithfulness of text generation.
Early work on document-grounded dialog used simple overlap based metrics like unigram F1 scores between the response and retrieved passages as a proxy for faithfulness \cite{dinan2018wizard,thulke2023dstc}.
\citet{fadeeva-etal-2024-fact} make use of uncertainty quantification to evaluate the factuality of generated responses.
Other work does not consider the claim decomposition step and directly verify the full response against the reference \cite{honovich-etal-2022-true-evaluating,juhasz2024responsibleretrievalaugmentedgeneration, kovács2025lettucedetecthallucinationdetectionframework}.

\section{Conclusion}

Ensuring faithfulness of LLM outputs is crucial for improving the reliability of climate-related RAG setups.
Our study evaluates automated faithfulness assessment methods.
According to our metric, recent LLMs like Llama 3.1 Instruct and GPT-4o provide much higher faithfulness than Llama 2 Chat or the climate-specific ClimateGPT model.
Based on our experiments, we assume that the main difference comes from the instruction fine-tuning and other post-training steps and not from the pre-training.

For ClimateGPT, we then do a detailed analysis, which subsets of the IFT data are most important for faithfulness.
We show that faithful closed-ended prompts in training also improve the faithfulness in the context of RAG and that it is crucial to avoid unfaithful training examples in the IFT data.
With these insights, we develop ClimateGPT Faithful+ which improves ClimateGPT's faithfulness from 30\% to 57\% according to our automatic metric.
These results are confirmed by additional experiments on an additional task as well as by using an additional metric to measure faithfulness.

These initial findings point to promising directions for future work.
Rather than discarding unfaithful training examples, one potential approach is to enrich them by retrieving supporting passages for each claim and using those passages as context during training.
For cases where no suitable evidence can be retrieved, synthetic context could be generated using a LLM.
This would keep a larger portion of the data while still encouraging faithful model behaviour.

\section*{Limitations}

In this work, we discuss results from our ongoing work towards more faithful LLMs for RAG on climate questions.
While our preliminary results are promising, there are still many open questions and limitations.

While RAGAs \cite{es-etal-2024-ragas} is a popular approach to evaluate faithfulness with RAG, we did not perform a systematic evaluation of its performance in the context of the task at hand.
Spot-checking of results during the development progress indicated that the metric is reliable enough for our purposes.
We tested the approach on two relevant climate datasets from the literature. The results are reported in \Cref{sec:cpr,sec:climate_fever}, but the results are inconclusive.
Thus, a more thorough human evaluation is needed to fully verify the adequacy of the metric for the task and to validate the improvements reported in this work.

The claim verification step in our pipeline currently focuses on verifying direct support via a given evidence passage.
This approach works well if the claim is directly expressed in the given passage and we can consider its content as truth.
In the context of evaluating faithfulness one can argue that this is a valid assumption.
But as soon as we want to also apply these methods to evaluate the factuality of more complex claims, this does not hold any more.
Often claims are not directly stated in a retrieved passage and more complex reasoning is required to identify the support.
Claims might express opinions or more holistic statements that require support from multiple sources to be considered as supported.
Also, a binary decision between supporting and not supporting might not be adequate in many cases, or more nuance is needed.
More complex claim verification approaches as proposed by \citet{Leippold2025climinator} partially address many of these points but are also much more complex and computationally expensive than the approach we use.

The behavior of a RAG system is highly dependent on the relevancy and adequacy of the retrieved passages.
In this work, we adopted the same knowledge base and retrieval method as used by \citet{thulke2024climategptaisynthesizinginterdisciplinary}.
Thus, our results are also limited to this specific setting and generalization to other settings needs to be studied.
Furthermore, the size of the knowledge base and the accuracy of the retrieval method limit the accuracy of the factuality evaluation during the evidence retrieval step.
Additional analysis would be needed to study the impact of these factors on the claim support wrt.\ the knowledge base, we consider as a proxy for factuality.

The ablation experiments on the IFT data focus on the climate-specific subsets.
We did not study the impact of the general domain IFT datasets included in IFT training, such as Open Assistant \cite{koepf2023openassistant}, Dolly\footnote{\url{https://huggingface.co/datasets/databricks/databricks-dolly-15k}} and FLAN v2 \cite{flan2023}.
Further, ClimateGPT is based on Llama 2.
In our experiments, we observed higher faithfulness for Llama 3.1 Instruct than for Llama 2 Chat.
The impact of the pre-training compared to different post-training steps on the faithfulness of the model remains unclear.

Finally, during our evaluation we only focused on claim support wrt.\ the reference and knowledge base which we consider as a proxy for faithfulness and factuality.
We do not consider additional quality factors like the helpfulness or adequacy of generated responses.
In some cases, a less faithful output can actually be more helpful or relevant.
For example the unfaithful parts in \Cref{fig:intro-example} like the information on the year might actually make the response more helpful for some users.

\bibliography{custom,anthology}

\appendix

\clearpage

\section{Full Results on the ClimateGPT IFT Task}
\label{appendix:full_results}

\Cref{tab:overall_results_all_models} shows the claim support of all models that we tested on the ClimateGPT IFT task.
In contrast to the table in the main part of the paper, here we also report the claim support wrt.\ the reference for the case that no RAG was used.
As the reference is not given to the model as additional input, we do expect low claim support.
The value is interesting as an indication for the percentage of claims that are faithful to the reference by chance.
Interestingly, we observe that the claim support of the original ClimateGPT models is close to this value.
This further supports the interpretation that these models do not make effective use of the provided context.
We omitted these results in the main part of the paper for better clarity as they are not directly relevant to the main claims of the paper.

\begin{table*}
    \centering
    \fontsize{10pt}{12pt}{
    \begin{tabular}{lrrcrrr}
        \toprule
                        &\textbf{\#Tokens}& \textbf{\#Parameters} &              & \textbf{Avg.}     & \multicolumn{2}{c}{\textbf{Claim Support wrt.}} \\
         \textbf{Model} &\textbf{in Trillion}& \textbf{in Billion}   & \textbf{RAG} & \textbf{\#Claims} & \textbf{Ref. \small{[\%]}} & \textbf{KB \small{[\%]}} \\
          \midrule
          GPT-4o & n/a & n/a & -  & 17.4 & 33                      & 68 \\
                 & & & \checkmark & 16.2 & \fontsize{11pt}{12pt}{\bf{72}}   & \fontsize{11pt}{12pt}{\bf{74}} \\
        \midrule
        LLama 3.1 Instruct &15& 8  & -  & 22.7 & 24                     & 59 \\
                           & &   & \checkmark & 17.3 & 67                     & \ul{72} \\
                            \cmidrule(rr){3-7}
                            & & 70 & -  & 21.8 & 25                     & 60 \\
                            &  &  & \checkmark & 16.1 & \ul{70}                     & \fontsize{11pt}{12pt}{\bf{74}} \\
        \midrule
            LLama 2 Chat & 2& 7  & -  & 23.3 & 25                    & 60 \\
                         &  &  & \checkmark & 21.2 & 48                   & 65 \\
                          \cmidrule(rr){3-7}
                          & &70 & -  & 25.1 & 24                     & 60 \\
                          & &   & \checkmark & 21.6 & 54                     & 68 \\
        \midrule
        ClimateGPT & 2& 7  & -  & 21.6 & 25                     & 59 \\
                   &  &  & \checkmark & 21.1 & 30                     & 61 \\
                    \cmidrule(rr){3-7}
                    & &70 & -  & 21.8 & 27                     & 61 \\
                    & &   & \checkmark & 22.2 & 30                     & 62 \\
        \midrule
        ClimateGPT Faithful+ (ours) & 2& 7 & -  & 20.2 & 27                     & 57 \\
                                  &  & & \checkmark & 19.2 & 57                     & 69 \\
        \bottomrule
    \end{tabular}}
    \caption{Results of all tested models for claim support wrt.\ the reference, as a metric of faithfulness, and wrt.\ the knowledge base (KB) as a metric for factuality for different large language models with and without RAG. The best values are in bold and the second best values underlined.}
    \label{tab:overall_results_all_models}
\end{table*}

\section{Evaluation Prompts}
\label{appendix:ragas_prompts}

\Cref{lst:prompt_claim_extraction} and \Cref{lst:prompt_claim_verification} show the prompts that were used for the claim extraction and verification steps in the evaluation pipeline.
Both prompts are based on the implementation of RAGAs\footnote{\url{https://github.com/explodinggradients/ragas}} \cite{es-etal-2024-ragas}.

\begin{lstlisting}[
        float=*,
        frame=single,
        breaklines=true,
        breakindent=0pt,
        framesep=3mm,
        numbers=none,
        basicstyle=\scriptsize\ttfamily,
        tabsize=2,
        caption={Prompt template used for Claim Extraction adapted from RAGAs.},
        label={lst:prompt_claim_extraction}
    ]
Given a question, an answer, and sentences from the answer, analyze the complexity of 
each sentence and break it down into one or more fully understandable statements.
Ensure that no pronouns are used in each statement and that every claim is explicit 
and self-contained. Format the output as a structured JSON response.

EXAMPLE
Question: Who was Albert Einstein and what is he best known for?
Answer: He was a German-born theoretical physicist, widely acknowledged to be one of 
the greatest and most influential physicists of all time. He was best known for 
developing the theory of relativity. He also made important contributions to the 
development of quantum mechanics.
Statements:
{
    "statements": [
        "Albert Einstein was a German-born theoretical physicist.",
        "Albert Einstein is recognized as one of the greatest and most influential physicists of all time.",
        "Albert Einstein was best known for developing the theory of relativity.",
        "Albert Einstein also made important contributions to the development of quantum mechanics."
    ]
}

YOUR TURN
Question: {{question}}
Answer: {{sentences}}
Statements:
\end{lstlisting}

\begin{lstlisting}[
    float=*,
    frame=single,
    breaklines=true,
    breakindent=0pt,
    framesep=3mm,
    numbers=none,
    basicstyle=\scriptsize\ttfamily,
    tabsize=2,
    caption={Prompt template used for Claim Verification adapted from RAGAs.},
    label={lst:prompt_claim_verification}
]
Your task is to judge the faithfulness of a series of claims based on a given context. For each claim you must return verdict as 1 if the claim can be directly inferred based on the context or 0 if the claim can not be directly inferred based on the context.

EXAMPLE 1:
Context: John is a student at XYZ University. He is pursuing a degree in Computer Science. He is enrolled in several courses this semester, including Data Structures, Algorithms, and Database Management. John is a diligent student and spends a significant amount of time studying and completing assignments. He often stays late in the library to work on his projects.

Claims:
1. John is majoring in Biology.
2. John is taking a course on Artificial Intelligence.
3. John is a dedicated student.
4. John has a part-time job.

Analysis:
{"analysis": [
{
    "claim": "John is majoring in Biology.",
    "reason": "John's major is explicitly mentioned as Computer Science. There is no information suggesting he is majoring in Biology.",
    "verdict": 0
},
{
    "claim": "John is taking a course on Artificial Intelligence.",
    "reason": "The context mentions the courses John is currently enrolled in, and Artificial Intelligence is not mentioned. Therefore, it cannot be deduced that John is taking a course on AI.",
    "verdict": 0
},
{
    "claim": "John is a dedicated student.",
    "reason": "The context states that he spends a significant amount of time studying and completing assignments. Additionally, it mentions that he often stays late in the library to work on his projects, which implies dedication.",
    "verdict": 1
},
{
    "claim": "John has a part-time job.",
    "reason": "There is no information given in the context about John having a part-time job.",
    "verdict": 0
}
]}

EXAMPLE 2:
Context: Photosynthesis is a process used by plants, algae, and certain bacteria to convert light energy into chemical energy.

Claims:
1. Albert Einstein was a genius.

Analysis:
{"analysis": [
{
    "claim": "Albert Einstein was a genius.",
    "reason": "The context and claim are unrelated.",
    "verdict": 0
}
]}

YOUR TURN:
Context: {{context}}
Claims:
{{claims}}
Analysis:
\end{lstlisting}

\section{Knowledge Base Details}
\label{appendix:retrieval_dataset}

\begin{table}[h]
    \begin{tabular}{lrr}
        \toprule
        \textbf{Source} & \textbf{\# Docs} & \textbf{\# 512 Chunks} \\ \midrule
        IPCC Reports & 16 & 17,897 \\
        Potsdam Papers & 390 & 8,539 \\
        Earth4All & 14 &  235 \\ 
        Other & 336 & 8,648 \\
        \midrule
        Total & 756 & 35,319 \\
        \bottomrule
    \end{tabular}
    \caption{Statistics of the different data sources of the ClimateGPT knowledge base.}
    \label{tab:retrieval_dataset}
\end{table}

\Cref{tab:retrieval_dataset} shows the statistics of the ClimateGPT knowledge base.

\section{RAG Prompts}
\label{appendix:rag_prompt}

\Cref{lst:rag_prompt} shows the prompt used in RAG for inference for all models except ClimateGPT.

\begin{lstlisting}[
    %float=*,
    frame=single,
    breaklines=true,
    breakindent=0pt,
    framesep=3mm,
    numbers=none,
    basicstyle=\scriptsize\ttfamily,
    linewidth=.95\linewidth,
    xleftmargin=.05\linewidth,
    tabsize=2,
    caption={Prompt used in RAG for inference for all models except ClimateGPT.},
    label={lst:rag_prompt}
]
You're a helpful assistant supporting users with their questions on climate change. Answer the question based on the given contexts. Make sure to only use information that is fully grounded in the contexts.

Context:
[[0]] "{passage[0].title}", {passage[0].year}
{passage[0].content}
{...}
[[4]] "{passage[4].title}", {passage[4].year}
{passage[4].content}

Question:
{question}
\end{lstlisting}

\section{Training Details}

In our training pipeline, we follow the setup from \citet{thulke2024climategptaisynthesizinginterdisciplinary}.
The models are trained using Megatron-LLM\footnote{\url{https://github.com/epfLLM/Megatron-LLM}} a fork of NVIDIA's Megatron-LM \footnote{\url{https://github.com/nvidia/megatron-lm}} by the EPFL LLM team.
A cosine learning rate schedule with a peak LR of $10^{-5}$ and 100 warmup steps are used.
The batch size is 64 and the sequence length is 4096. Additionally, a weight decay of $10^{-2}$ and dropout are used.

All 7B parameter models are trained with full parameter fine-tuning on 4xA100 80GB GPUs.
One training run takes approximately 4 hours, so in total 64 GPU hours were needed to train the models reported in this paper.

\section{CPR's RAG Dataset Evaluation}
\label{sec:cpr}

To further evaluate the generalization of ClimateGPT 7B Faithful+ to other datasets, we tested it on a set of question–passage pairs published by the Climate Policy Radar team \cite{juhasz2024responsibleretrievalaugmentedgeneration}. This dataset contains 1,013 examples, with the retrieved passages taken from Climate Policy Radar’s internal database. We generated responses using both ClimateGPT 7B and ClimateGPT 7B Faithful+, and evaluated their faithfulness to the provided reference passages using our RAGAs-based metric.
On this dataset, ClimateGPT 7B Faithful+ achieved a claim support of 58\%, substantially outperforming the base ClimateGPT 7B model, which achieved 44\%. These results demonstrate that the improvements made in the refined model generalize effectively to other climate-domain datasets.

In addition, \citet{juhasz2024responsibleretrievalaugmentedgeneration} also collected expert annotations for model outputs from GPT-4o, GPT-3.5, Gemini 1.0 and 1.5, and Mistral 7B v0.2. Each response was evaluated for faithfulness using a definition closely aligned with ours. Expert annotators labeled responses as either faithful (58.9\%), not faithful (9.6\%), not applicable (28\%), or don’t know (3.5\%). We used this data to evaluate how well our RAGAs-based metric aligns with human judgments. 
For the analysis, we focused only on examples that were labeled as either faithful or not faithful, excluding cases where the model refused to answer. This resulted in a total of 1,367 samples. 
To convert the claim support from our metric into a binary label for each example, we classify an output as faithful if the claim support exceeds 50\%. On this test set, our metric achieved an overall agreement of 86.7\% with the human annotations. However, accuracy varied between label categories: it reached 93.7\% for human-labeled faithful responses, but only 29.5\% for not faithful ones. Notably, \citet{juhasz2024responsibleretrievalaugmentedgeneration} themself acknowledged that their annotations were sometimes \enquote{too noisy along the faithfulness dimension}. In addition, limited spot-checking on our part more frequently agreed with our metric's assessments than with the human annotations.

\section{Evaluation with LettuceDetect}
\label{appendix:results_lettuce_detect}

In addition to our primary faithfulness evaluation using RAGAs, we include results using \textbf{LettuceDetect} \cite{kovács2025lettucedetecthallucinationdetectionframework}, a recent hallucination detection framework designed for RAG systems. LettuceDetect is a token-level classifier based on ModernBERT \cite{warner2024smarterbetterfasterlonger}, trained on the RAGTruth dataset \cite{niu-etal-2024-ragtruth} to identify hallucinated spans in LLM responses given the input question and context. As LettuceDetect’s definition of hallucination closely aligns with our notion of faithfulness, we use it to validate the results obtained with RAGAs.

For our evaluation, we convert LettuceDetect’s span-level predictions into a binary faithfulness score by marking a generation as faithful if no hallucinated spans are detected. Results, using the \texttt{lettucedetect-large-v1} variant of the model, are reported in \Cref{tab:lettuce_detect_results}.

The LettuceDetect results support the conclusions drawn from our RAGAs-based evaluation. Without RAG, both ClimateGPT and ClimateGPT Faithful+ achieve low scores (6\% and 2\% respectively), providing a baseline for how often generations align with the reference context by chance. With RAG, ClimateGPT Faithful+ shows a substantial improvement, reaching 34\% hallucination-free responses compared to only 6\% for the original ClimateGPT. This underpins the claim support results obtained with RAGAs (57\% vs.\ 30\%), reinforcing the effectiveness of our instruction fine-tuning strategy in improving the model’s ability to ground its generations in the retrieved context.

\begin{table*}
    \centering
    \fontsize{10pt}{12pt}{
    \begin{tabular}{lrrrr}
        \toprule
                        & \textbf{\#Tokens} & \textbf{\#Parameters} & \multicolumn{2}{c}{\textbf{Hallucination-Free Responses [\%]}} \\
         \textbf{Model} & \textbf{in Trillion}& \textbf{in Billion}   & \textbf{w/o RAG} & \textbf{w/ RAG} \\
        \midrule
        GPT-4o                    &n/a& n/a   & 9  & \textbf{73} \\
        \midrule
        LLaMA 3.1 Instruct        &15& 8  & 4  & \underline{65} \\
                                  && 70 & 2  & \underline{65} \\
        \midrule
        LLaMA 2 Chat              &2& 7  & 2  & 18 \\
                                  && 70 & 2  & 29 \\
        \midrule
        ClimateGPT               &2& 7  & 4  & 6 \\
                                 && 70 & 4  & 4 \\
        \midrule
        ClimateGPT Faithful+ (ours) &2& 7 & 2  & 34 \\
        \bottomrule
    \end{tabular}}
    \caption{Percentage of hallucination-free responses according to \texttt{lettucedetect-large-v1} \cite{kovács2025lettucedetecthallucinationdetectionframework} for various models with and without RAG. The best values are in bold and the second best values underlined.}
    \label{tab:lettuce_detect_results}
\end{table*}

\section{ClimateGPT IFT Example Output}

\Cref{fig:ocean-example} shows an example output of ClimateGPT 7B and the ClimateGPT 7B Faithful+ model proposed in this work.
While both responses address the question in the prompt and are generally correct, the refined model’s response demonstrates a significantly closer alignment with the reference passages.

\begin{figure*}[t]
    \begin{minipage}{\textwidth}
        \begin{mdframed}
            \fontsize{8pt}{9.6pt}\selectfont
        \textbf{Prompt}

        How do ocean ecosystems, ranging from coral reefs to the deep sea, support biodiversity and provide essential services? Answer in a single detailed paragraph.

        \textbf{Context}
        
        [0] "Understanding the Effectiveness of Coastal Nature-based Solutions: Practitioner-based Learning", 2023\\
        despite the high cost and failure rate. oyster reefs are effective for wave attenuation, and research suggests that focusing on positive species interactions can provide a framework for restoration. biodiversity enhancement supports a shellfish reef's ability to provide ecosystem services. ecosystem services research highlights that healthy ecosystems provide high - quality services, while stressed ecosystems produce degraded services and may harm human well - being. most ecosystem services are supported by biodiversity. biodiversity in coastal systems contributes to providing ecosystem services such as fish habitat, nutrient cycling, and various cultural services. additionally, biodiversity may play a role in the\newline
        [1] "The Role of Blue Carbon in Climate Change Mitigation and Carbon Stock Conservation", 2021\\
        is an essential to sustainably manage and develop marine resources to their maximum potential. environmental measures should tackle both terrestrial and marine ecosystems, with one as a continuum of the other. coral reef restoration can increase coastal resilience to sea level rise and flooding and provide valuable environmental services for local populations. water pollution in rivers contributes to ocean ecosystem degradation, via eutrophication and the formation of \newline
        [2] "The Ocean and Cryosphere in a Changing Climate", 2020\\
        level of risk than the high emission scenario. changing marine ecosystem services and human well - being ecosystem services are environmental processes and functions that provide benefits to people and support human well - being. marine ecosystem services are generated throughout the ocean, from shallow water to the deep sea. these services can be divided into provisioning services, regulating services, supporting services, and cultural services. they are also described as'nature's contribution to people '. changes in key ecosystem\newline
        [3] "Climate Change 2022: Impacts,Adaptation and Vulnerability", 2022\\
        logical shifts and trophic mismatches, changes in communities and biodiversity, time of emergence of climate - impact signals in ecological systems from background natural variability, and changes in biomass, primary productivity, and carbon export. coastal ecosystems and seas warm - water coral reefs warm - water coral reef ecosystems, which house a significant portion of marine biodiversity, provide food, income, and shoreline protection to coastal communities. these ecosystems are threatened by climate - induced and non - climate drivers, including ocean warming, marine heatwaves, ocean acidification, sea - level rise, tropical cyclones \newline
        [4] "Understanding the Effectiveness of Coastal Nature-based Solutions: Practitioner-based Learning", 2023 \\
        research indicates that biodiversity plays a crucial role in the functioning of ecosystem services provided by habitats. recent studies have focused on the climate mitigation capacity of forests in the context of biodiversity and nature - based solutions ( nbs ). challenges arise in prioritizing monoculture plantations for carbon storage over intact and diverse ecosystems. the identity and richness of species, as well as the concept of'niche complementarity ', contribute to boosting ecosystem function and their services. significant restoration efforts are underway for shellfish reefs, despite the high cost and failure rate. oyster reefs" \newline
        
        \textbf{ClimateGPT 7B (Faithfulness: 19\%, Factuality: 67\%)}  

        Ocean ecosystems, encompassing a diverse range from coral reefs to the deep sea, \textcolor{rot}{demonstrate unparalleled biodiversity} and provide a multitude of essential services.         
        \textcolor{rot}{In the coral reef ecosystem, vibrant arrays of colorful fish and intricate coral structures serve as vital breeding grounds and habitats for countless marine organisms}.
        \textcolor{rot}{These ecosystems act as natural breakwaters, reducing coastal erosion and protecting shorelines}.
        Additionally, they \textcolor{rot}{store vast amounts of carbon, contributing to climate change mitigation}.         
        Moving out from the coast, the deep sea presents a different world, \textcolor{rot}{characterized by complete darkness, extreme pressure, and cold temperatures}.
        Despite the harsh conditions, \textcolor{rot}{it sustains unique species} adapted to these environments, \textcolor{rot}{many of which are still undiscovered}.
        \textcolor{rot}{The deep sea plays a crucial role in regulating climate and weather patterns}, driven by its \textcolor{rot}{immense biological pump, which captures carbon from the surface} and \textcolor{rot}{transports it to the deep oceans}.         
        In essence, ocean ecosystems are not only a source of \textcolor{rot}{aesthetic beauty} and \textcolor{rot}{recreation} but also \textcolor{rot}{linchpins of the Earth's ecological balance} and \textcolor{rot}{a treasure trove of biodiversity}.

        \textbf{ClimateGPT 7B Faithful+ (Faithfulness: 92\%, Factuality: 96\%)}

        Ocean ecosystems, including coral reefs and the deep sea, \textcolor{rot}{play a vital role in biodiversity support} and providing essential services.
        They provide provisional services, regulating services, supporting services, and cultural services.
        Coral reefs contribute to coastal resilience to sea level rise and flooding and provide valuable environmental services for local populations.
        However, they are threatened by climate-induced and non-climate drivers, \textcolor{rot}{impacting coastal ecosystems} and warm-water coral reefs. These ecosystems provide food, income, and shoreline protection to coastal communities.
        Restoration efforts, such as oyster reefs, are essential for wave attenuation, but their failure and high cost are challenges. Recent research has focused on the climate mitigation capacity of forests in the context of biodiversity and nature-based solutions. Prioritizing monoculture plantations for carbon storage over intact and diverse ecosystems is a challenge.
        The identity and richness of species and the concept of niche complementarity contribute to boosting ecosystem function and services.
        \end{mdframed}
    \end{minipage}
    \caption{Example comparing the outputs of ClimateGPT and ClimateGPT 7B Faithful+ on one example of the held-out test set. Parts marked in red correspond to claim that are not supported in the given context according to our automatic evaluation.}
    \label{fig:ocean-example}
\end{figure*}

\begin{figure*}[t]
    \begin{minipage}{\textwidth}
        \begin{mdframed}
            \fontsize{8pt}{9.6pt}\selectfont
        \textbf{Prompt}

        What is the purpose of the Global Stocktake?

        \textbf{Context}
        
        [0] Recalling Article 14 of the Paris Agreement, the objective of the Global Stocktake is to assess the collective progress towards achieving the purpose of the Agreement and its long-term goals, considering mitigation, adaptation and the means of implementation and support, in light of equity and the best available science. The outcome of the global stocktake shall inform Parties in updating and enhancing, in a nationally determined manner, their actions and support in accordance with the relevant provisions of this Agreement, as well as in enhancing international cooperation for climate action. The GST two-year process, begins with an information collection and preparation component, followed by a technical assessment component and culminates in a consideration of outputs component, which includes the outcomes of the GST. The SBSTA and the SBI invited submissions on views on the approach to the consideration of outputs component of the first global stocktake. This submission responds to that invitation and sets out to identify important characteristics for the outputs phase and the shaping of the outcomes of the GST. \newline
        [1] · Paragraph 34 of 19/CMA.1 decides that the outputs of this component of the global stocktake should: a) Identify opportunities for and challenges in enhancing action and support for collective progress in relation to the thematic areas of the global stocktake referred to in paragraph 6(b) above, as well as possible measures and good practices and international cooperation and related good practices;\newline
        [2] The SBSTA and the SBI invited submissions on views on the approach to the consideration of outputs component of the first global stocktake. This submission responds to that invitation and sets out to identify important characteristics for the outputs phase and the shaping of the outcomes of the GST. Approach to the Consideration of Outputs:
        
        \textbf{ClimateGPT 7B}  

        The Global Stocktake is a key component of the Paris Agreement, \textcolor{rot}{scheduled to begin in 2023}. It involves a comprehensive assessment of global progress in addressing climate change, \textcolor{rot}{considering scientific evidence, economic implications, and societal impacts}. The Stocktake aims to inform the updates to countries' climate commitments, \textcolor{rot}{known as \enquote{nationally determined contributions} (NDCs)}, \textcolor{rot}{to ensure they are in line with the agreement's objectives and targets}. This process seeks to enhance international cooperation and commitment to mitigating climate change on a global scale.

        \textbf{ClimateGPT 7B Faithful+}

        The Global Stocktake's purpose, as outlined in Article 14 of the Paris Agreement, is to assess the collective progress made towards achieving the Agreement's purpose and long-term goals, which encompass mitigation, adaptation, and means of implementation. This assessment is conducted considering equity and the best available science. The Global Stocktake process spans two years, beginning with information collection and preparation and culminating in a consideration of outputs. The outputs of the latter component are decided upon in paragraph 34 of 19/CMA.1, which states that they should identify opportunities for enhancing action and support for collective progress in relation to thematic areas of the global stocktake and possible measures and good practices.
        \end{mdframed}
    \end{minipage}
    \caption{Example comparing the outputs of ClimateGPT and ClimateGPT 7B Faithful+ on one example from the Climate Policy Radar data. Parts marked in red correspond to claim that are not supported in the given context according to our automatic evaluation.}
    \label{fig:stocktake-example}
\end{figure*}

\section{Climate FEVER Claim Verification}
\label{sec:climate_fever}

To evaluate our claim verification method, we applied it to the Climate-FEVER dataset \cite{diggelmann2021climatefeverdatasetverificationrealworld}.
The dataset consists of 1,535 claims, each paired with 5 corresponding evidence passages. Each claim-evidence pair is annotated by at least two annotators as either supported, refuted, disputed, or lacking sufficient information. For our analysis, we classify refuted and lacking sufficient information as not supported, and exclude all pairs labelled as disputed.
In addition, we only include examples where all annotators agree on the label, resulting in a total of 1,146 claims and 3,348 claim-evidence pairs.
On this subset, the RAGAs-based claim verifier achieves an overall accuracy of 67.1\%. For pairs with the gold label \enquote{not supported}, the accuracy is 99.7\%, while for supported pairs, it drops to 20.6\%.
Qualitatively, we observe that our claim verifier is relatively strict, requiring the claim to be explicitly stated in the evidence.
An example of this is given in \Cref{lst:climate_fever_example}.
In this instance, the evidence does not explicitly state that polar bears are one of the affected species. However, it could be argued that it is plausible to infer that polar bears are included among the \enquote{many species} mentioned in the evidence.

\begin{lstlisting}[
    float,floatplacement=H,
    frame=single,
    breaklines=true,
    breakindent=0pt,
    framesep=3mm,
    numbers=none,
    basicstyle=\scriptsize\ttfamily,
    linewidth=.95\linewidth,
    xleftmargin=.05\linewidth,
    tabsize=2,
    caption={Climate-FEVER example where our claim verifier disagrees with the gold label.},
    label={lst:climate_fever_example}
]
Claim:
Global warming is driving polar bears toward extinction

Evidence:
[Global Warming] Environmental impacts include the extinction or relocation of many species as their ecosystems change, most immediately the environments of coral reefs, mountains, and the Arctic.

Gold label: supported
Predicted label: not_supported
\end{lstlisting}

\end{document}